\documentclass[journal]{IEEEtran}
\IEEEoverridecommandlockouts

\usepackage[ruled,linesnumbered]{algorithm2e}
\usepackage{cite}
\usepackage{amsmath,amssymb,amsfonts}
\usepackage{algorithmic}
\usepackage{graphicx}
\usepackage{textcomp}
\usepackage{xcolor}
\usepackage{enumerate}
\usepackage{booktabs}
\usepackage{diagbox}
\usepackage{supertabular}
\usepackage{caption}
\usepackage{float} 
\usepackage{verbatim}
\usepackage{bm} 
\usepackage{breqn}
\usepackage{stfloats}
\usepackage{amsthm}
\usepackage{hyperref}
\usepackage{makecell}
\usepackage{amssymb}
\usepackage{subfigure}
\usepackage{multirow}
\usepackage{array}
 \usepackage{booktabs}
 \newtheorem{definition}{Definition}

\makeatletter

\newcommand{\Rmnum}[1]{\expandafter\@slowromancap\romannumeral #1@}
\makeatother

\def\BibTeX{{\rm B\kern-.05em{\sc i\kern-.025em b}\kern-.08em
    T\kern-.1667em\lower.7ex\hbox{E}\kern-.125emX}}

\begin{document}
\title{Hybrid RAG-empowered Multi-modal LLM for Secure Data Management in Internet of  Medical Things: A Diffusion-based Contract Approach}

\author{Cheng Su, Jinbo Wen, Jiawen Kang*, Yonghua Wang, Yuanjia Su,  Hudan Pan, \\
Zishao Zhong, and M. Shamim Hossain

\thanks{Cheng Su is with the School of Automation and the Guangdong-HongKong-Macao Joint Laboratory for Smart Discrete Manufacturing, Guangdong University of Technology, Guangzhou 510006, China (e-mail: chengsu9251@163.com).}
\thanks{Jinbo Wen is  with the College of Computer Science and Technology, Nanjing University of Aeronautics and Astronautics, Nanjing 210016, China (e-mail: jinbo1608@163.com).}
\thanks{Jiawen Kang is with the School of Automation and the Guangdong Basic Research Center of Excellence for Ecological Security and Green Development, Key Laboratory for City Cluster Environmental Safety and Green Development of the Ministry of Education, Guangdong University of Technology, Guangzhou 510006, China (e-mail: kavinkang@gdut.edu.cn).}
\thanks{Yonghua Wang is with the School of Automation and the Key Laboratory of Intelligent Detection and IoT in Manufacturing, Ministry of Education, Guangdong University of Technology, Guangzhou 510006, China (e-mail: wangyonghua@gdut.edu.cn).}
\thanks{Jiayuan Su is with the School of Automation and the Key Laboratory of Intelligent Information Processing and System Integration of IoT, Ministry of Education, Guangdong University of Technology, Guangzhou 510006, China (e-mail: syj1216902331@163.com).}
\thanks{Hudan Pan is with Guangzhou University of Traditional Chinese Medicine, State Key Laboratory of Traditional Chinese Medicine Syndrome, The Second Affiliated Hospital of Guangzhou University of Chinese Medicine, Chinese Medicine Guangdong Laboratory, Hengqin, 519000, China (e-mail: panhudan1988@163.com).}
\thanks{Zishao Zhong is with the Second Affiliated Hospital of Guangzhou University of Chinese Medicine, Chinese Medicine Guangdong Laboratory, Hengqin, 519000, China (e-mail: zhongzishao@gzucm.edu.cn).}
\thanks{M. Shamim Hossain is with the Department of Software Engineering, College of Computer and Information Sciences, King Saud University, Riyadh 12372, Saudi Arabia (e-mail: mshossain@ksu.edu.sa).}
\thanks{(\textit{*Corresponding author: Jiawen Kang})}
}

\maketitle

\begin{abstract}
Secure data management and effective data sharing have become paramount in the rapidly evolving healthcare landscape, especially with the growing integration of the Internet of Medical Things (IoMT). The rise of generative artificial intelligence has further elevated Multi-modal Large Language Models (MLLMs) as essential tools for managing and optimizing healthcare data in IoMT. MLLMs can support multi-modal inputs and generate diverse types of content by leveraging large-scale training on vast amounts of multi-modal data. However, critical challenges persist in developing medical MLLMs, including security and freshness issues of healthcare data, affecting the output quality of MLLMs. To this end, in this paper, we propose a hybrid Retrieval-Augmented Generation (RAG)-empowered medical MLLM framework for healthcare data management. This framework leverages a hierarchical cross-chain architecture to facilitate secure data training. Moreover, it enhances the output quality of MLLMs through hybrid RAG, which employs multi-modal metrics to filter various unimodal RAG results and incorporates these retrieval results as additional inputs to MLLMs. Additionally, we employ age of information to indirectly evaluate the data freshness impact of MLLMs and utilize contract theory to incentivize healthcare data holders to share their fresh data, mitigating information asymmetry during data sharing. Finally, we utilize a generative diffusion model-based deep reinforcement learning algorithm to identify the optimal contract for efficient data sharing. Numerical results demonstrate the effectiveness of the proposed schemes, which achieve secure and efficient healthcare data management.

\end{abstract}


\IEEEpeerreviewmaketitle

\section{Introduction}
\label{sec:introduction}

The healthcare system has seen rapid advancements with the integration of advanced technologies like cloud computing, the Internet of Things (IoT), and Artificial Intelligence (AI). These innovations have transformed the sector, giving rise to the Internet of Medical Things (IoMT), which is an interconnected network of medical devices and applications that collect and transmit vital healthcare data\cite{huang2023internet}. IoMT has not only paved the way for more intelligent and efficient healthcare systems, but also catalyzed the generation, storage, and analysis of vast amounts of big healthcare data\cite{isgut2022systematic}, including omics data, clinical records, electronic health records, etc\cite{gupta2023perspective}. Although the exponential growth in healthcare data volume holds the potential to revolutionize the healthcare industry by providing insights into patient care, disease patterns, and treatment effectiveness, it also requires sophisticated tools for its analysis and interpretation. Fortunately, Generative Artificial Intelligence (GenAI) as a new branch of AI has emerged as a potent technology within IoT landscape\cite{li2024filling, tang2024digital}, enabling the effective analysis of vast datasets and generation of diverse content\cite{wen2024generative,chen2024generative}. In particular, GenAI enhances data management by analyzing complex patient records and treatment data, enabling more efficient sharing of critical healthcare information\cite{bisht2023efficient}.

Large Language Models (LLMs), as a technological application of GenAI, can achieve general-purpose language generation and conventional natural language processing tasks, which hold the potential to significantly transform healthcare data management in IoMT\cite{yuan2024efficient}. With the integration of multi-modal data into LLMs, patients can effectively comprehend many aspects of their physical health through Multi-modal LLMs (MLLMs)\cite{mesko2023impact}. For example, the latest GPT-4, equipped with vision capabilities and exceptional performance in natural language processing tasks, can be fine-tuned as a powerful guidance tool in the healthcare domain\cite{achiam2023gpt}. However, the sheer size and complexity of MLLMs necessitate efficient retrieval mechanisms to enhance their performance further. Retrieval-Augmented Generation (RAG) is a cutting-edge technique that boosts the reliability and accuracy of GenAI models by retrieving facts from an external knowledge base\cite{lewis2020retrieval}. Furthermore, RAG can capitalize on the similarity between the alignment vectors of the query to retrieve pertinent data, thereby enhancing user prompts by integrating relevantly retrieved data within the context, enabling MLLMs to generate accurate and contextually appropriate responses\cite{wen2024generative}. Thanks to the prominent capabilities of RAG, the integration of MLLMs and RAG has been widely used in various domains\cite{gao2023retrieval, 10531073}.

Despite the advancements in RAG-empowered MLLMs, there are several persistent challenges in the application of these technologies for healthcare data management in IoMT: 1) Since healthcare data is normally multi-modal and stored in different databases in a distributed manner, unimodal RAG using a single search manner, such as vector similarity search and keyword search\cite{10531073}, may not efficiently retrieve multi-modal healthcare data to support LLM tasks that handle multiple modes. 2) The application of MLLMs in analyzing healthcare data poses significant security risks and privacy concerns\cite{ullah2023scalable}. Healthcare data is highly sensitive, and any breach or misuse can have severe consequences for patients and healthcare providers\cite{isgut2022systematic}. Thus, ensuring the confidentiality and integrity of healthcare data during MLLM processing is a critical concern. 3) Pre-trained medical MLLMs can result in inaccurate inferences during task-specific fine-tuning due to biases in the dataset. Hence, incorporating fresh high-quality healthcare data is crucial for fine-tuning MLLMs to avoid incorrect learning patterns\cite{wen2024generative}. 4) Considering the problem of information asymmetry, healthcare data holders often have more data information, and appropriate incentive mechanisms need to be implemented to encourage healthcare data holders to provide accurate and up-to-date information, which is helpful to enhance the medical diagnostic quality of MLLMs empowered by RAG.

To address these challenges, we propose a hybrid RAG-empowered medical MLLM framework for healthcare data management in IoMT. Specifically, we allow participants to share data without the involvement of a central institution by implementing cross-chain techniques, which support secure and efficient data or asset transfers across multiple chains, effectively mitigating single-point-of-failure risks and enhancing overall security\cite{Kangblockchain}. To enhance the diagnostic quality of MLLMs, we leverage hybrid multi-modal RAG to further refine the retrieval results. Compared with RAG-empowered LLMs, we employ multi-modal metrics to filter multiple unimodal RAG results and incorporate these retrieval results into MLLMs as additional inputs. Furthermore, we apply Age of Information (AoI) to indirectly evaluate the quality of healthcare data and utilize a contract theory model to encourage participants to share fresh data, thus coping with the information asymmetry of data sharing. Besides, considering the dynamic environment of data sharing, we use Generative Diffusion Model (GDM)-based Deep Reinforcement Learning (DRL) algorithms to efficiently find the optimal contract\cite{du2024diffusion}. The key contributions of this paper are summarized as follows:  

\begin{itemize}
    \item We develop a novel hybrid RAG-empowered MLLM framework for healthcare data management in IoMT. This framework facilitates secure interactions between healthcare data holders and the MLLM service provider using a cross-chain system for secure healthcare data transmission, and MLLMs can improve their quality and complete specific tasks by employing hybrid RAG to retrieve multi-modal healthcare data. 
    \item To optimize time-sensitive learning tasks within MLLM services, we apply AoI as a data freshness metric to indirectly assess the quality of healthcare data. Furthermore, we formulate a contract theory model to incentivize healthcare data holders to contribute high-quality healthcare data with small AoI, thus improving the inference performance of hybrid RAG-empowered MLLMs.
    \item To tackle the high-dimensional complexity of the formulated problem, we employ GDM-based DRL algorithms to determine the optimal contract for efficient data sharing. Numerical results demonstrate the effectiveness of the proposed GDM-based scheme, showing a $20.35\%$ performance improvement over DRL-based schemes, highlighting its superiority in this paper.
\end{itemize}

The remainder of this paper is organized as follows. Section \ref{sec:Related Work} reviews the related work. In Section \ref{sec:System Model}, we propose a hybrid RAG-empowered medical MLLM framework based on cross-chain technology to enhance data management in IoMT. In Section \ref{sec:problem}, we introduce a contract theory model to motivate healthcare data holders to provide high-quality healthcare data. In Section \ref{sec:Diffusion}, we present GDM-based DRL algorithms for optimal contract design. Section \ref{sec:numerical results} provides a performance analysis of the proposed schemes. Finally, Section \ref{sec:Conclusion} concludes this paper. The main notations in our article are summarized in Table \ref{Paper_notation}.

\section{Related Work}
\label{sec:Related Work}

\subsection{RAG-empowered LLMs}

RAG has incredible capabilities in enhancing the accuracy and reliability of LLM output by incorporating additional information sources, such as external knowledge bases, and augmenting user prompts with relevant retrieval data in context\cite{lewis2020retrieval}. As a novel technique, RAG  allows LLMs to bypass retraining, allowing access to the most up-to-date information to generate reliable output through retrieval-based generation\cite{gao2023retrieval}. In \cite{lewis2020retrieval}, the authors introduced RAG, demonstrating its ability to improve the accuracy and relevance of generated text by incorporating retrieved documents into the generation process. The authors in \cite{omrani2024hybrid} proposed a hybrid RAG method that integrates Sentence-Window and Parent-Child approaches and demonstrated that the proposed method outperforms current state-of-the-art RAG techniques. The authors in \cite{wen2024generative} introduced a carbon emission optimization framework that integrates RAG and LLMs, making a significant impact on GenAI efforts to reduce carbon emissions. Moreover, RAG is gradually emerging as a promising tool for healthcare applications, for example, it can optimize the interpretation of clinical guidelines for liver disease with the help of external medical knowledge\cite{kresevic2024optimization}. In addition, the authors in \cite{yuan2023ramm} retrieved similar image-text pairs based on image-text contrast similarity, and utilized the retrieval attention module to blend the representation of images and questions with the retrieved images and texts, demonstrating effectiveness in simple biomedical visual question answering. 

\subsection{LLMs for Data Management}

IoMT has significantly improved healthcare data management by enabling the seamless collection, transmission, and analysis of vast amounts of patient data through interconnected devices\cite{bisht2023efficient}. LLMs further enhance this process by offering advanced capabilities in processing unstructured data, extracting valuable insights, and supporting decision-making with greater efficiency and accuracy\cite{yuan2024efficient}. Recent advancements in LLMs have significantly contributed to data management, with numerous research efforts focusing on various aspects such as data analysis, predictive modeling, and decision support systems. Some studies use the strong interpretative abilities of LLMs as agents to continuously improve data storage, data analysis, and additional areas\cite{achiam2023gpt, chen2022program}. For instance, the authors in \cite{achiam2023gpt} introduced GPT-4, which has demonstrated remarkable capabilities in understanding and generating human-like text, facilitating various data management tasks\cite{achiam2023gpt}. The authors in \cite{zhou2024db} presented an LLM-based database framework that leverages LLMs for automatic prompt generation and model fine-tuning, which performs highly effective in query rewriting and index tuning\cite{zhou2024db}. In \cite{zhang2024data}, the authors introduced Data-Copilot, which is a data analysis agent capable of autonomously querying, processing, and visualizing vast amounts of data to meet various human needs. Existing works mainly focus on unimodal LLMs processing text data, while relatively insufficient research has been done on the integration of multi-modal data such as text, images, and structured data. Addressing this gap could substantially boost the functionality of data management systems, especially in complex and data-intensive IoMT.

\subsection{Contract Theory for Data Sharing}

Contract theory is a branch of economics that studies how contractual arrangements can be designed to align incentives between parties with asymmetric information\cite{hou2017incentive}, and it has been widely used in wireless communication, AI, and other fields\cite{hou2017incentive, Kangblockchain}. In the context of data sharing, information asymmetry often arises because data holders possess more information about the data than data users. Contract theory can effectively incentivize data sharing by ensuring that both parties benefit from the exchange\cite{Kangblockchain}. For example, the authors in \cite{lim2020dynamic} proposed a two-period incentive mechanism for healthcare applications, which takes into account the Willingness To Participate (WTP) of users and satisfies intertemporal incentive compatibility. This dynamic contract design meets essential constraints and achieves higher profits compared to a uniform pricing scheme. In the context of a mobile AI-generated content network with Unmanned Aerial Vehicles (UAVs), the authors in \cite{wen2023freshness} proposed an AoI-based contract theory model to incentivize the contribution of fresh data between UAVs. The authors in \cite{kang2019incentive} proposed an effective incentive mechanism. This mechanism integrates reputation and contract theory to motivate high-reputation mobile devices with high-quality data to engage in model learning in a federated learning scenario. In addition to the above work, several efforts have been made to develop contract theory models under prospect theory to facilitate user-centric sensing data sharing\cite{Kangblockchain}.

Despite these advancements, the application of diffusion-based contract theory for data sharing remains unexplored. While existing contract theory models are valuable, they often fall short of addressing the intricate challenges associated with the dynamic and multifaceted nature of data sharing. In many real-world contexts, data sharing is not a simple transaction but a process spanning multiple stages and involving diverse participants\cite{wen2024diffusion}. Diffusion-based contract theory incorporates the spread and evolution of information over time and space, enabling a deeper understanding of incentives and behaviors among data holders and users\cite{du2024diffusion}. By accounting for diffusion patterns, this approach allows for more accurate assessments of uncertainties and risks tied to data sharing. Consequently, diffusion-based contract models enable the design of adaptive and effective contractual arrangements that support efficient and sustainable data sharing.

\section{Hybrid RAG-empowered Medical MLLM Framework}\label{sec:System Model}
\begin{table}[t]
\renewcommand{\arraystretch}{1.6}
\captionsetup{font = small}
\caption{Key Mathematical Notations of this Paper}\label{Paper_notation}
\begin{tabular}{m{0.8cm}|m{6.8cm}} 
\toprule[1.5pt]
\hline
\multicolumn{1}{c|}{\textbf{Notation}}  & \multicolumn{1}{c}{\textbf{Definition}} \\ \hline
$t_{trans}$ & The transmission time of healthcare data   \\ \hline
$t_u$ & The time of completing consensus among blockchains   \\ \hline
$\ell$ & The size of healthcare data   \\ \hline
$\tau$ & The transmission rate of healthcare data between the health center and hospitals \\ \hline
$\overline{A}_{m}$ & The average AoI for data sharing by data holder $m$ \\ \hline
$\overline{A}_{max}$ & The maximum permissible value for the AoI. \\ \hline
$R_k$ & Reward to the type-$k$ healthcare data holders for the MLLM service provider  \\ \hline
$\delta_k$ &  The $k$-th type healthcare data holder \\ \hline
$f_k$ &  The update frequency of the type-$k$ healthcare data holder  \\ \hline
$\alpha$ & Overall zero-shot accuracy of MLLMs   \\ \hline
$S_k$ & The satisfaction function of the MLLM service provider obtained from the type-$k$ healthcare data holder   \\ \hline
$\beta$ & The unit profit associated with type-$k$ healthcare data holder  \\ \hline
$Q_k$ & The proportion of type-$k$ healthcare data holder in healthcare industry \\ \hline
$\boldsymbol{\pi}_\omega$ & Contract design policy with parameters $\omega$\\ \hline
$\boldsymbol{\epsilon}_\omega$ & Contract generation network with parameters $\omega$\\ \hline
$q_\varphi$ & Contract quality network with parameters $\varphi$\\\hline
$\boldsymbol{\pi}_{\omega '}$ & Target contract design policy with parameters $\omega '$\\\hline
$\boldsymbol\epsilon '_{\omega '}$ & Target contract generation network with parameters $\omega '$\\\hline
$q_{\varphi'}'$ & Target contract quality network with parameters $\varphi '$\\\hline
$\Psi^{0}$ & Optimal contract design \\\hline
\bottomrule[1.5pt]
\end{tabular}
\end{table}

In this section, we propose a hybrid RAG-empowered medical MLLM framework in IoMT. The detailed methodologies for cross-chain interaction in MLLM training and the utilization of hybrid RAG-empowered MLLM agents for data management are discussed in the following subsections.

\subsection{Cross-Chain Interaction in MLLM Training}
In the health center, robust aggregate MLLMs are developed by training on vast amounts of high-quality multi-modal healthcare data\cite{yuan2023ramm}. Due to privacy concerns and factors such as patient willingness and incentive structures, hospitals may be reluctant to upload all healthcare data to a central health center\cite{Kangblockchain}. Additionally, the significant computational power required for training MLLMs, combined with the lack of high-performance computing resources at most hospitals, limits the feasibility of using federated learning in this framework\cite{xu2024cached, yang2023detfed}. These constraints necessitate alternative approaches for managing data and training models effectively.

In response to these multifaceted challenges, blockchain and cross-chain technologies have emerged as powerful solutions to facilitate secure and decentralized data sharing across healthcare networks. Blockchain technology ensures data integrity and transparency by providing an immutable ledger, while cross-chain technology enables seamless interoperability between diverse blockchain networks, allowing secure data and asset transactions across chains\cite{bisht2023efficient}. Recent studies further demonstrate the potential of cross-chain frameworks in significantly enhancing data security and enabling effective collaboration across distributed healthcare systems\cite{Kangblockchain, bisht2023efficient}. To this end, we incorporate cross-chain technology to enable hospitals to securely upload sensitive healthcare data and conduct secure transactions with the health center, ensuring both data privacy and system efficiency.

\begin{figure*}[t]
\centering
\includegraphics[width=0.95\textwidth]{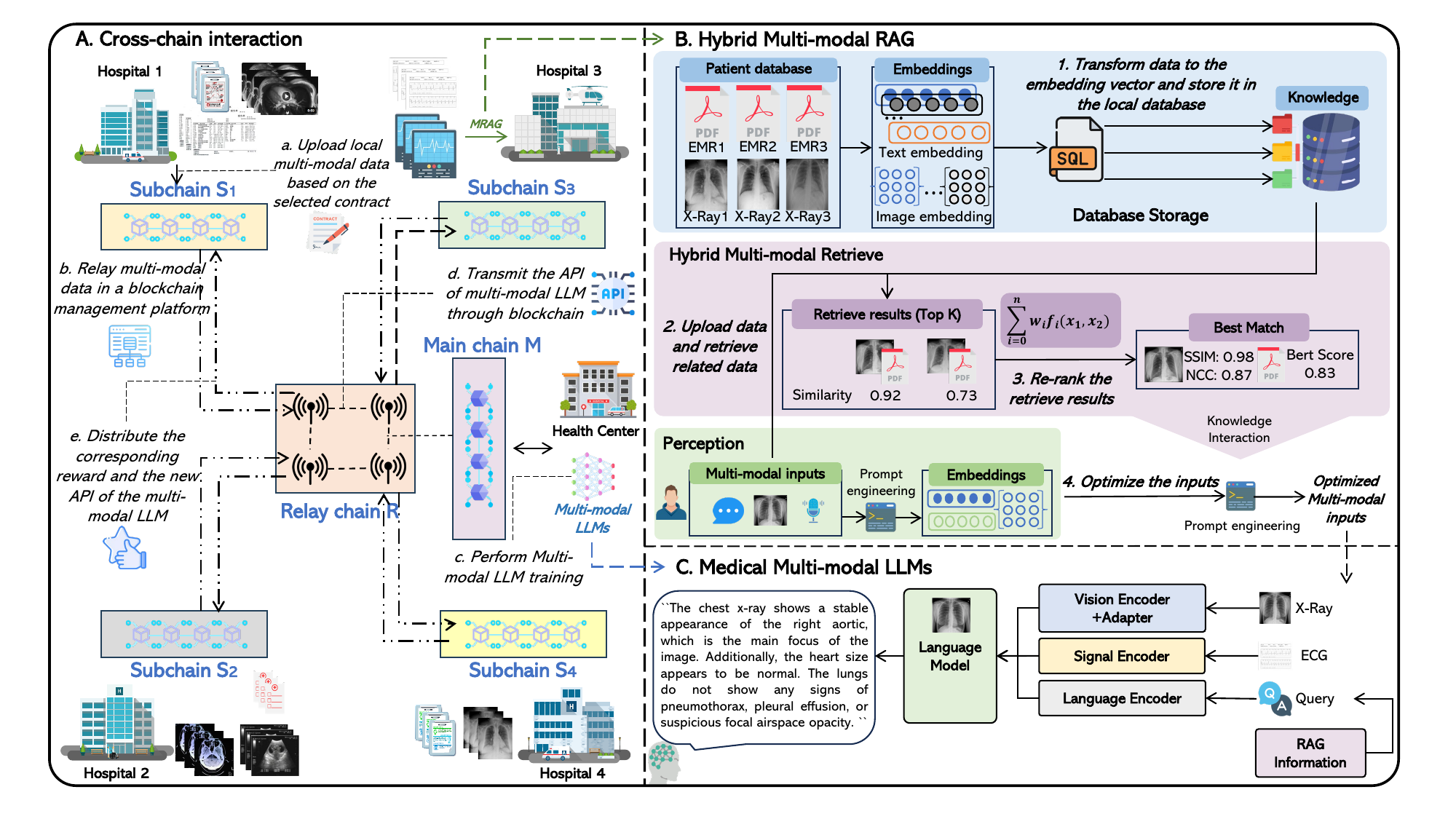}
\captionsetup{font = small}
\caption{The overview of the hybrid RAG-empowered medical MLLM framework for healthcare data management in IoMT. Part A shows the cross-chain interaction for secure healthcare data sharing. Part B depicts the processes of multi-modal input optimization based on a hybrid multi-modal RAG module. Part C presents the framework of MLLM inference based on the multi-modal healthcare data.}\label{System figure}
\end{figure*}

As shown in Fig. \ref{System figure}, a robust health center utilizes a main chain to manage the comprehensive collection of healthcare data and model updates, and multiple subchains are employed to handle specific tasks from hospitals in diverse regions. These subchains utilize IoMT devices to collect real-time healthcare data from patients, such as temperature, heart rate, and blood pressure, enabling doctors to develop personalized treatment strategies based on a comprehensive analysis of multiple patient attributes\cite{huang2023internet, wen2024generative}. In addition to data collection, the subchains also manage MLLM configurations, and other workflow tasks to support effective healthcare delivery. The main chain ensures centralized oversight, while the sub-chains enable efficient and secure data management operations. Specifically, the main chain $M$ sends data collection tasks to the relay chain $R$. When the subchains $S_1$, $S_2$, $S_3$, and $S_4$ receive the tasks, hospitals will upload local multi-modal healthcare data based on the selected contracts (Step \textit{a} in Fig. \ref{System figure}). Upon successful verification of cross-chain requests by the miners of the relay chain $R$, the relay chain $R$ returns a readiness confirmation, allowing subchains to upload multi-modal healthcare data (Step \textit{b} in Fig. \ref{System figure}) \cite{Kangblockchain}. After transmitting the multi-modal healthcare data to the main chain $M$, the health center initiates MLLM training (Step \textit{c} in Fig. \ref{System figure}). Once MLLM training is completed, the health center and hospitals can access the MLLM Application Programming Interface (API) or the weights file of MLLMs through the relay chain $R$ (Step \textit{d} in Fig. \ref{System figure}), and the health center rewards them with monetary compensation based on their data contribution (Step \textit{e} in Fig. \ref{System figure})\cite{Kangblockchain}.

\subsection{Hybrid RAG-empowered MLLM Framework for Data Management in IoMT}
Data management tasks in hospitals and the health center include data storage, analysis, and retrieval. When multi-modal healthcare data is gathered into the subchains, MLLMs categorize the data by type and store it in the appropriate databases. As healthcare data is needed, MLLMs retrieve and analyze the data, thus meeting the specific requirements of data management tasks\cite{zhou2024db}. To further enhance the ability of MLLMs to analyze multi-modal healthcare data, we design a hybrid multi-modal RAG module\cite{omrani2024hybrid}, which is integrated with a data sharing mechanism inspired by contract theory, ensuring that the MLLM data analysis is conducted with high quality and strong privacy protection, allowing secure and effective handling of multi-modal healthcare data. As shown in the right of Fig. \ref{System figure}, the workflow of hybrid RAG-empowered MLLMs for data management in IoMT is presented as follows:

\textit{Step 1. Store multi-modal healthcare data:} Hospitals and the health center collect available multi-modal healthcare data, convert them into vectors specific to each modality using an embedding model, and store these vectors in the local knowledge database with Structured Query Language (SQL) tools\cite{lewis2020retrieval}.

\textit{Step 2. Retrieve multi-modal healthcare data:} When a task query is received, the hybrid multi-modal RAG system uses the same embedding model from \textit{Step 1} to convert the query into a vector. Then, the system calculates similarity scores between the task query vector and vectors within the knowledge database, retrieving and prioritizing the top $K$ vectors that most closely match the task query\cite{gao2023retrieval}.

\textit{Step 3. Re-rank the retrieved information:} When all relevant information, particularly multi-modal data, is fed directly into MLLMs, it can lead to information overload and may reduce attention to critical details due to the inclusion of irrelevant content\cite{gao2023retrieval}. To address this, the system further screens the results by applying our Multi-modal Information Similarity (MIS) metric, which is calculated by
\begin{equation}
\label{MIS}
    \begin{split}
        MIS=\sum_{i=0}^nw_if_i(x_1,x_2),
    \end{split}
\end{equation}
where $f_i(\cdot)$ represents the similarity measure function between the task query and the source data in the database and is determined freely according to the requirements of the specific task, $x_1$ and $x_2$ are the unimodal data corresponding to the task query and the source data in the database, respectively, and the weight factor $w_i$ is used to characterize the proportion of each the similarity measure function $f_i(\cdot)$. When the results are re-ranked and filtered by MIS, the retrieved optimized healthcare information is then used to expand the context in the prompt.

\textit{Step 4. Optimize the multi-modal inputs:} Upon completing the retrieval process, we employ prompt engineering based on zero-shot prompting technology to optimize and synthesize a coherent prompt that integrates the original multi-modal task query with the retrieved healthcare data\cite{gao2023retrieval, liu2023optimizing}. By using `probability' as a control keyword, this refined prompt enhances the credibility of the entire prompt and improves the MLLM's ability to generate accurate and contextually relevant responses.

\textit{Step 5. Generate the corresponding content based on the inputs:} Upon receiving the multi-modal inputs, MLLMs connect each modal input to its respective pre-trained encoder model, where a pre-trained linear projection adapter is employed to unify all processed embeddings\cite{yuan2023ramm, li2024llava}. This linear projection adapter, trained on 600K image-text pairs from PMC-15M, standardizes the embeddings, allowing pre-trained LLMs to generate the corresponding content based on the inputs\cite{li2024llava, gao2023retrieval}.

In the generation of MLLMs, hybrid RAG enhances the generation quality by effectively incorporating relevant information from various sources. However, RAG does not improve the generalization ability of MLLMs, indicating that the ability of MLLMs to apply learned information to new and unseen contexts remains limited, limiting its overall learning ability. To improve the output quality of MLLMs, an essential way is to continuously incorporate new healthcare data for training. Thus, we propose an incentive mechanism to encourage data holders to share updated healthcare data.

\section{Problem Formulation}\label{sec:problem} 
In this section, we begin by developing a metric for healthcare data quality, followed by the formulation of utility functions for both healthcare data holders and the MLLM service provider. Finally, we formulate a contract theory model to motivate healthcare data holders to contribute high-quality healthcare data.

The training of MLLMs relies heavily on a large volume of high-quality data\cite{achiam2023gpt}. Unfortunately, most healthcare data are stored in hospital databases in various regions. Without data sharing, these valuable resources remain untapped, hindering MLLM development. Furthermore, the effectiveness of MLLMs is directly influenced by the quality of the data used in training. Therefore, it is critical to implement an incentive mechanism that encourages hospitals to share healthcare data. Referring to \cite{Kangblockchain}, we consider there are multiple hospitals in diverse regions and a health center as an example. The health center acts as the MLLM service provider, and the hospitals in diverse regions serve as the healthcare data holders, represented by a set of $\mathcal{M}=\{1,\ldots,m,\ldots,M\}$. Initially, we propose a healthcare data quality metric through the AoI metric to assess the quality of healthcare data utilized for fine-tuning MLLMs. Subsequently, acting as the data task publisher, the MLLM service provider employs a contract theory model to encourage $M$ healthcare data holders to engage in data sharing \cite{wen2023freshness}.

\subsection{Healthcare Data Quality Metrics}
AoI has gained broad acceptance as a metric for assessing data freshness, especially within wireless communication networks \cite{lim2020information}. In this paper, AoI is described as the duration between the data gathering at the hospital and the finalization of MLLM training. Lower AoI correlates with higher-quality MLLM output for healthcare applications. As described in \cite{wen2023freshness}, we propose a healthcare data quality metric through AoI, which is relevant for scenarios involving periodic data updates.

To generalize, we define the size of healthcare data as $\ell$ (bytes) and the transmission rate between the health center and hospitals as $\tau$ (bytes per second). Hence, the transmission time of healthcare data is $t_{trans} = \ell / \tau$ \cite{wen2023freshness}. Meanwhile, we denote $t_u$ as the time of completing a consensus process among blockchains\cite{Kangblockchain}. Therefore, we represent the length of a single time slot $t$ as $t = t_{trans} + t_u$ \cite{Kangblockchain}. To maintain data freshness, each healthcare data provider $m$ periodically updates its healthcare data, with $\theta_{m}$ indicating the length of a single time slot in each update cycle. The refreshment of healthcare data happens in the initial time slot of the cycle. Referring to \cite{zhang2018towards}, the AoI for a data request made in the $i$-th time slot is $(i+1)t$ for $i = 2,\ldots, \theta_m - 1$, and for requests initiated in the first or last time slot, the AoI is $2t$. Due to the Poisson process\cite{wen2023freshness, Kangblockchain}, data requests are equally likely to occur in any time slot, with a probability of $ 1 / \theta_m$. Therefore, the average AoI for data sharing by healthcare data holder $m$ is given by
\begin{equation}
    \label{average_AOI}
    \begin{split}
        \overline{A}_{m}(\theta_{m})=\frac{2}{\theta_{m}}(2t)+\sum_{i=2}^{\theta_{m-1}}\frac{(i+1)t}{\theta_{m}}=t\bigg(\frac{1}{\theta_{m}}+\frac{\theta_{m}}{2}+\frac{1}{2}\bigg).
    \end{split}
\end{equation}

Recognizing that a large AoI can degrade the quality of sensing data, we define the healthcare data quality metric $G(A_m)$ based on AoI as
\begin{equation}
    \begin{split}
        G(\overline{A}_m)=\overline{A}_{max}/\overline{A}_m,
    \end{split}
\end{equation}
where $\overline{A}_{max}$ represents the maximum permissible value for the AoI. The healthcare data quality metric plays a critical role in the quality of MLLM services. Given that (\ref{average_AOI}) is a convex function in relation to the update cycle $\theta_m$, increasing $\theta_m$ results in a decrease in AoI\cite{wen2023freshness}. Consequently, there is a tradeoff in managing AoI, which can be optimized by modifying the update cycle.

\subsection{Healthcare Data Holder Utility}
In the context of healthcare data sharing for MLLM services, the utility for each healthcare data holder $m$ is the difference between the reward $R_m$ and the cost $C_m$ incurred by data sharing tasks, expressed as $U_m = R_m - C_m$ \cite{Kangblockchain}. According to \cite{zhou2021towards}, the cost for healthcare data holder $m$ is defined as $C_m = \xi_mf_m$ \cite{wen2023freshness}, with $f_m=\frac1{\theta_m}$ representing the update frequency and $\xi_m$ denoting the cost of each update \cite{Kangblockchain}. Thus, the utility of healthcare data holder $m$ is
\begin{equation}
    \begin{split}
        U_m = R_m - \xi_mf_m.  \newline
    \end{split}
\end{equation}

Due to information asymmetry, the MLLM service provider lacks precise knowledge of the update cost of each healthcare data holder. To address this, the MLLM service provider classifies data holders into discrete types by using statistical distributions derived from historical data, and its expected utility will be optimized\cite{Kangblockchain}. By classifying $M$ healthcare data holders into various types, we denote the $k$-th type healthcare data holder as $\delta_k = 1/\xi_k$ and group them into a set $\mathcal{K}=\{\delta_{k}:1\leq n\leq K\}$, where a smaller update cost corresponds to a higher healthcare data holder type, and the healthcare data holder types are organized as $\delta_{1}\leq\delta_{2}\leq\cdots\leq\delta_{K}$. Thus, the utility of the type-$k$ healthcare data holder is given by
\begin{equation}
    \begin{split}
        U_k(R_k, f_k) = R_k - \frac{f_k}{ \delta_k}.  \newline
    \end{split}
\end{equation}

\begin{table*}[htbp]
\renewcommand{\arraystretch}{1.5}
 \centering
 \tiny
 \captionsetup{font = small}
 \caption{The Parameters of Accuracy in Different Domains of LLaVA-Med-10K/60K.}
 \setlength{\tabcolsep}{10pt}
 
 {\fontsize{10}{11}\selectfont
  \begin{tabular}{ccccccccc}
  \toprule[1.5pt]
  \hline
    \multirow{2}{*}{\textbf{Model}} &\multicolumn{2}{c}{\textbf{Question Types}}    & \multicolumn{5}{c}{\textbf{Domains}} & \multirow{2}{*}{\textbf{Overall}}
    \\\cmidrule(lr){2-3} \cmidrule(lr){4-8}
    & Conversation & Description  & CXR & MRI & Histology & Gross & CT \\
    (Question Count) & (143) & (50) & (37) & (38) & (44) & (34) & (40) & (193) \\\midrule
    LLaVA-Med-10K  & 42.4 & 32.5 & 46.1 & 36.7 & 43.5 & 34.7 & 37.5 & 39.9 \\
    LLaVA-Med-60K  & 53.7 & 36.9 & 57.3 & 39.8 & 49.8 & 47.4 & 52.4 & 49.4
    \\\hline
    \bottomrule[1.5pt]
 \end{tabular}
 \selectfont
 }
 \label{The Parameters of Accuracy}
\end{table*}

\subsection{MLLM Service Provider Utility}
Due to the quality of MLLMs' output being affected by healthcare data freshness, large AoI leads to poor output for MLLMs and reduces the satisfaction of the MLLM service provider. Referring to \cite{xu2024cached}, The satisfaction function for the MLLM service provider, based on type-$k$ healthcare data holders, is defined as
\begin{equation}
    \begin{split}
        S_k = \alpha \log(G(\overline{A}_k) + 1),    
    \end{split}
\end{equation}
where $\alpha$ is the overall zero-shot accuracy of MLLMs for various services. For example, the zero-shot accuracy of LLaVA-Med as a medical LLM across different domains is presented in Table \ref{The Parameters of Accuracy}. Here, the value of $\alpha$ is determined by past experience when applied to various services\cite{li2024llava}.

Owing to information asymmetry, the MLLM service provider just knows the total count and type distributions of healthcare data holders, without detailed information about the type of each healthcare data holder\cite{wen2023freshness, Kangblockchain}. Thus, the expected utility of the MLLM service provider is calculated in the following manner\cite{zhou2021towards, Kangblockchain}
\begin{equation}\label{expected utility}
    \begin{split}
        U_s(\boldsymbol{f},\boldsymbol{R})=\sum_{k=1}^K Q_k(\beta S_k-R_k).    
    \end{split}
\end{equation}
Here, $\beta > 0$ represents the unit profit associated with $S_k$, while $Q_k$ is the probability that a healthcare data holder is type-$k$, subject to the constraint that the sum of these probabilities equals $1$, i.e., $\sum_{k=1}^KQ_k=1$. Additionally, $\boldsymbol{R}=[R_k]_{1\times K}$ and $\boldsymbol{f}=[f_k]_{1\times K}$ represent the vectors of rewards and update frequencies for all $K$ types of healthcare data holders, respectively.

\subsection{Contract Formulation}
To prevent rational healthcare data holders from supplying low-quality data in pursuit of higher rewards, a robust method is required to maintain MLLM service quality\cite{wen2023freshness}. Given that contract theory is an economic tool for effectively designing incentive mechanisms under conditions of asymmetric information, we propose a contract theory model for the MLLM service provider. This model leverages contract theory to effectively motivate healthcare data holders to provide timely data updates, ensuring the reliability of MLLM services\cite{Kangblockchain}.

In this scenario, the MLLM service provider takes the lead in designing a set of contract items and offers them to $K$ healthcare data holders. Based on its type, each healthcare data holder selects the most appropriate contract item, denoted by  $\Psi_k=\{(f_k,R_k), k\in\mathcal{K}\}$, where $f_k$ represents the update frequency for type-$k$ healthcare data holders, and $R_k$ is the reward given to type-$k$ healthcare data holders as an incentive for its contribution. To guarantee that each healthcare data holder opts for the most advantageous contract item for its type, the designed contract must adhere to both Incentive Compatibility (IC) and Individual Rationality (IR) constraints.

\begin{definition}
(IR) The contract item for a type-$k$ healthcare data holder guarantees a non-negative utility, formulated as
\end{definition}
\begin{equation}\label{IR}
    \begin{split}
        R_{k} - \frac{  f_k }{ \delta_{k} }  \geq 0, \:\forall k \in \mathcal{K}.
    \end{split}
\end{equation}
\begin{definition}
(IC) A healthcare data holder of type-$k$ will choose the contract item  $(f_k,R_k)$ tailored to its type rather than any other contract item $(f_{i}, R_{i}),\: i\in \mathcal{K}$, and $i\neq k$, i.e.,
\end{definition}
\begin{equation}
    \begin{split}\label{IC}
        R_{k} - \frac{  f_k }{ \delta_{k} }  \geq  R_{i} - \frac{  f_i }{ \gamma_{k} }  , \:\forall k, i \in \mathcal{K},\: k\neq i.
    \end{split}
\end{equation}

To maximize the expected utility of the MLLM service provider, the optimization problem can be formulated as
\begin{equation}\label{problem1}
    \begin{split}
        & \max_{ \bm{f}, \bm{R}} U_s(\boldsymbol{f},\boldsymbol{R})=\sum_{k=1}^K Q_k(\beta S_k-R_k)  \\
        &\:\: \text{s.t.}  \:\: R_{k} - \frac{  f_k }{ \delta_{k} }  \geq 0, \:\forall k \in \mathcal{K},\\
        & \qquad R_{k} - \frac{  f_k }{ \delta_{k} }  \geq  R_{i} - \frac{  f_i }{ \gamma_{k} }  , \:\forall k, i \in \mathcal{K},\: k\neq i,\\
        & \qquad f_k \geq 0, R_k \geq 0, \delta_k > 0,\: \forall k \in \mathcal{K}.
    \end{split}
\end{equation}

Traditional mathematical solutions often struggle to effectively adapt to the complexity and dynamic changes inherent in data sharing environments\cite{ye2021joint}. In response, we leverage GDMs, a key component of GenAI, which excel not only in image generation but also in optimizing network performance\cite{su2024privacy, wen2024generative}. Building on similar approaches\cite{du2023ai, wen2024diffusion}, we employ GDMs as a more efficient solution for identifying optimal contracts. This approach capitalizes on the generative capabilities of GDMs to capture uncertainties and fluctuations in network conditions, allowing for more accurate identification of optimal contracts in real-time scenarios and effectively addressing the high-dimensional and intricate nature of the problem\cite{du2024diffusion}.

\section{Generative Diffusion-based Contract Design}\label{sec:Diffusion}
In this section, we initially formulate the contract design between the MLLM service provider and healthcare data holders as a Markov Decision Process (MDP). Then, we present a GDM-based contract generation model to determine the optimal contract.

\subsection{MDP Formulation}
\subsubsection{State space}
To find the optimal contract item, i.e., $(f_{k,}^*,R_k^*)$, $\begin{matrix}k&\in&\mathcal{K} \end{matrix}$, the system first adds Gaussian noise to the initial contract sample. In the current diffusion round  $t=1,2,\ldots,T$, the state space affecting the optimal contract design is defined as
\begin{equation}\label{state}
    \begin{split}
        \boldsymbol{s}\triangleq\{M,K,\overline{A}_{max},\mathcal{Q},\mathcal{K}\},
    \end{split} 
\end{equation}
where $M$ and $K$ are constant values, while $\overline{A}_{max}$, $\mathcal{Q}= (Q_1,\ldots,Q_K)$, and $\mathcal{K} = (\delta_1,\ldots,\delta_K)$ are generated randomly in the current diffusion round $t$.

\subsubsection{Action space}
As the MLLM service provider designs a contract $\Psi$ to motivate healthcare data holders to provide high-quality healthcare data, the action $\boldsymbol{a}^t$ at round $t$ is defined as
\begin{equation}
    \begin{split}
        \boldsymbol{a}^{t}\triangleq\{\Psi^{t}\},
    \end{split}
\end{equation}
where $\Psi^{t}=\{(f_{k}^{t},R_{k}^{t}),k\in\mathcal{K}\}$ determines the update frequency and reward for type-$k$ healthcare data holders.

\subsubsection{Immediate reward}
Following the action $\boldsymbol{a}^t$, the MLLM service provider achieves an immediate reward $r(\boldsymbol{s}, \boldsymbol{a}^t)$ aimed at maximizing the expected utility described in (\ref{expected utility}) while ensuring compliance with the IR (\ref{IR}) and IC (\ref{IC}) constraints. Thus, the reward function is defined as
\begin{equation}\label{reward_s}
    \begin{split}
        r(\boldsymbol{s},\boldsymbol{a}^{t})= 
        \begin{cases}
            U_s^{t}(\boldsymbol{f},\boldsymbol{R}), & \text{if $\boldsymbol{a}^{t}$ satisfies (\ref{IR}) and (\ref{IC}),} \\
            U_p, & \text{otherwise,}
        \end{cases}
    \end{split} 
\end{equation}  
where $U_s^{t}(\boldsymbol{f}, \boldsymbol{R})$ denotes the expected utility of the MLLM service provider during round $t$, and $U_p\leq 0$ serves as the penalty for violating either the IR or IC constraints.

\begin{algorithm}[t]
\DontPrintSemicolon
\SetAlgoLined
    \caption{GDM-based Optimal Contract Design for Data Sharing}\label{AI_Contract}

    \KwIn{GDM's hyperparameters, e.g.,  diffusion step $T$, discount factor $\gamma$, and exploration noise $\varepsilon$.}
    \KwOut{The optimal contract design $\boldsymbol{a}^0$.}
    \#\#\#\#\#\# \textit{Phase 1: Initialization}\\
    Initialize replay buffer $\mathcal{D}$, contract generation network $\boldsymbol{\epsilon}_\omega$, contract quality network $q_\varphi$, target contract generation network $\boldsymbol{\epsilon}'_{\omega'}$, target contract quality network $q'_{\varphi'}$.\\
    
    \#\#\#\#\#\# \textit{Phase 2: Training} \\
    \For{\rm{Episode} $e=1$ to $E_{max}$}
    {
        Initialize a random process $\mathcal{N}$ to facilitate contract design exploration. \\
        \For{\rm{Step} $z=1$ to $Z_{max}$}
        {
            Observe the current environment $\boldsymbol{s}_z$.\\
            Set $\boldsymbol{a}_z^T$ as Gaussian noise and generate contract design $\boldsymbol{a}_z^0$ by denoising $\boldsymbol{a}_z^T$ based on  (\ref{denoise}). \\
            Execute contract design $\boldsymbol{a}_z^0$ and observe the reward $r_z$ (\ref{reward_s}). \\
            Store record $(\boldsymbol{s}_z,\boldsymbol{a}_z^0,r_z, \boldsymbol{s}_{z+1})$ into replay buffer $\mathcal{D}$. \\
            Sample a random mini-batch of $N$ records $(\boldsymbol{s}_i,\boldsymbol{a}_i^0,r_i, \boldsymbol{s}_{i+1})$ from replay buffer $\mathcal{D}$. \\
            Update the contract quality network by minimizing (\ref{Q_update}).\\
            Update the contract generation network by computing the policy gradient (\ref{policy_update}).\\
            Update the target networks:
            $\omega'\leftarrow\eta\omega+(1-\eta)\omega'$,
            $\varphi'\leftarrow\eta \varphi+(1-\eta)\varphi'$. \\
        }
    }
    \textbf{return} The trained contract generation network $\boldsymbol{\epsilon}_\omega$. \\
    \#\#\#\#\#\# \textit{Phase 3: Inference} \\
    Input the environment vector $\boldsymbol{s}$ (\ref{state}). \\
    Generate the optimal contract design $\boldsymbol{a}^0$ based on (\ref{denoise}). \\
    \textbf{return} $\boldsymbol{a}^{0}=\{(f_{k}^*,R_{k}^*),k\in\mathcal{K}\}$.
\end{algorithm}

\subsection{GDMs for Optimal Contract Design}
Compared with DRL algorithms that directly optimize model parameters\cite{tang2023digital}, GDMs can enhance contract design through an iterative process of denoising the initial distribution\cite{du2023ai, wen2024diffusion}. The diffusion model network maps the environmental state to contract design, which constitutes the contract design policy represented as $\boldsymbol{\pi}_\omega(\boldsymbol{a}|\boldsymbol{s})$ with parameters $\omega$. The policy $\boldsymbol{\pi}_\omega(\boldsymbol{a}|\boldsymbol{s})$ designed to generate an optimal contract over multiple time steps can be expressed as
\begin{equation}
    \begin{split}
        \begin{aligned}
            \boldsymbol{\pi}_{\omega}(\boldsymbol{a}|\boldsymbol{s})& =p_{\omega}(\boldsymbol{a}^{0},\ldots,\boldsymbol{a}^{T}|\boldsymbol{s})  \\
            &=\mathcal{N}(\boldsymbol{a}^T;\mathbf{0},\mathbf{I})\prod_{t=1}^Tp_\omega(\boldsymbol{a}^{t-1}|\boldsymbol{a}^t,\boldsymbol{s}^t),
        \end{aligned}
    \end{split}
\end{equation}
Here, $\boldsymbol{\pi}_{\omega}(\cdot)$ represents the reverse process of the conditional diffusion model and $p_\omega(\boldsymbol{a}^{t-1}|\boldsymbol{a}^t,\boldsymbol{s}^t)$ is modeled as a Gaussian distribution $\mathcal{N}(\boldsymbol{a}^{t-1};\boldsymbol{\mu}_\omega(\boldsymbol{a}^t,\boldsymbol{s},t),\boldsymbol{\Sigma}_\omega(\boldsymbol{a}^t,\boldsymbol{s},t))$, where the covariance matrix $\boldsymbol{\Sigma}_\omega(\boldsymbol{a}^t,\boldsymbol{s},t)$ is formulated as\cite{du2023ai}
\begin{equation}
    \begin{split}
        \boldsymbol{\Sigma}_\omega(\boldsymbol{a}^t,\boldsymbol{s},t) = \delta_t\mathbf{I},
    \end{split}
\end{equation}
where $\delta_t\in(0,1)$ is a hyperparameter determined before model training, and $\mathbf{I}$ is the identity matrix. Consequently, the mean $\boldsymbol{\mu}_\omega(\boldsymbol{a}^t,\boldsymbol{s},t)$ can be given by\cite{du2023ai}
\begin{equation}
    \begin{split}
        \boldsymbol{\mu}_\omega(\boldsymbol{a}^t,\boldsymbol{s},t) = \frac1{\sqrt{\chi_t}}\bigg(\boldsymbol{a}^t-\frac{\delta_t}{\sqrt{1-\bar{\chi_t}}}\boldsymbol{\epsilon}_\omega(\boldsymbol{a}^t,\boldsymbol{s},t)\bigg),
    \end{split}
\end{equation}
where $\chi_t=1-\delta_t$, $\bar{\chi_{t}}=\prod_{i=0}^{t}\delta_{j}$, and $\boldsymbol{\epsilon}_\omega$ denotes  the contract generation network. We first sample $\boldsymbol{a}^T \sim \mathcal N (\textbf{0}, \textbf{I})$ and then sample from the reverse diffusion chain parameterized by $\omega$, which is given by\cite{du2023ai}
\begin{equation}\label{denoise}
    \boldsymbol{a}^{t-1}|\boldsymbol{a}^t=\frac{\boldsymbol{a}^t}{\sqrt{\chi_t}}-\frac{\delta_t}{\sqrt{\chi_t(1-\bar{\chi_t})}}\boldsymbol\epsilon_\omega(\boldsymbol{a}^t,\boldsymbol{s},t)+\sqrt{\delta_t}\boldsymbol\epsilon.
\end{equation}

Referring to \cite{du2023ai, wen2024diffusion}, we effectively train the contract design policy $\boldsymbol{\pi}_{\omega}$ to enhance the training quality of the contract generation network $\boldsymbol{\epsilon}_\omega$. Additionally, inspired by the concept of the Q-function \cite{van2016deep}, we introduce a contract quality network $q_\varphi(\boldsymbol{s},\Psi)$. The training of the contract quality network utilizes the double Q-learning technique to minimize the Bellman operator, involving two critic networks $q_{\varphi_1}, q_{\varphi_2}$ and the corresponding target critic networks $q_{\varphi_1^{\prime}}, q_{\varphi_2^{\prime}}$. We define $q_\varphi = \min\{q_{\varphi_1}, q_{\varphi_2}\}$, and the optimal contract design policy that maximizes the expected cumulative utility of the client is expressed as\cite{wen2024diffusion}

\begin{equation}\label{policy_update}
    \boldsymbol{\pi} = \arg\max_{\boldsymbol{\pi}_{{\omega}}} \Bbb E \Bigg[\sum_{z = 0} ^ Z \gamma^z (r(\boldsymbol{s}_z,\boldsymbol{a}_z)-\varsigma\boldsymbol{\pi}_{\omega}(\boldsymbol{s}_z)\log\boldsymbol{\pi}_{\omega}(\boldsymbol{s}_z))\Bigg],
\end{equation}
where $\gamma$ represents the discount factor, $\boldsymbol{a}_z$ represents the action in the training step $z$, and $\varsigma$ represents the temperature coefficient controlling the strength of the entropy. 

We define the target policy as $\boldsymbol{\pi}_{\omega '}$, and the optimization of $\varphi_{i}$ for $i=1,2$ is performed by minimizing the following objective function\cite{wen2024diffusion}:

\begin{equation}\label{Q_update}
\begin{split}
    &\Bbb E_{(\boldsymbol{s}_z, \boldsymbol{a}_z, \boldsymbol{s}_{z+1}, r_z)\sim \mathcal{B}_z}\Big[\sum_{l=1,2}(r(\boldsymbol{s}_z,\boldsymbol{a}_z)-q_{\varphi_l}(\boldsymbol{s}_z,\boldsymbol{a}_z)\\
    &\quad\quad\quad+\gamma^z(1-d_{z+1})\boldsymbol{\pi}_{\omega '}(\bm{s}_{z+1})q'_{\varphi'}(\bm{s}_{z+1}))^2\Big],
\end{split}
\end{equation}
where $\mathcal{B}_z$ is a mini-batch of transitions sampled from the experience replay memory $\mathcal{D}$ in the training step $z$ and $d_{z+1}$ is a $0$-$1$ variable denoting the terminated flag.

The pseudo-code of the proposed GDM-based contract generation scheme is shown in Algorithm \ref{AI_Contract}, which consists of three phases, and its computational complexity is $\mathcal{O}(|\omega|+|\varphi|+E_{max}Z_{max}(T|\omega| + |\varphi|))$. In the proposed GDM-based contract generation scheme, denoising techniques are employed to generate optimal contract designs \cite{du2023ai, wen2024diffusion}. By integrating exploration noise into the contract design and executing it, the process accumulates exploration experience, contributing to the enhancement of contract quality.

\begin{figure*}[t]
\centering
\includegraphics[width=0.95\textwidth]{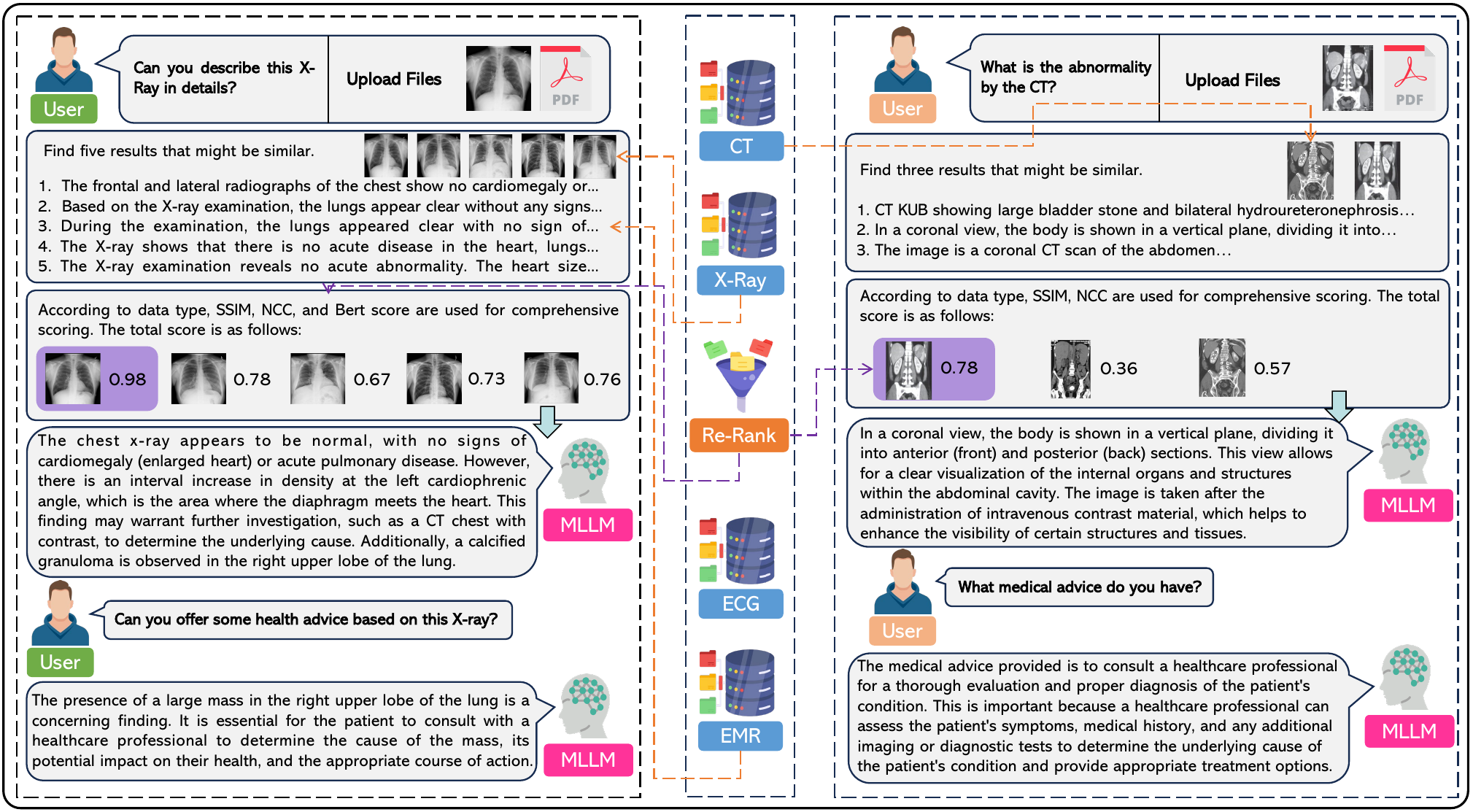}
\captionsetup{font = small}
\caption{A real case study of hybrid RAG-empowered medical MLLMs. In the proposed hybrid RAG-empowered medical MLLM, the RAG initially retrieves healthcare data using unimodal methods. Next, we re-rank the information using metrics such as structural similarity index measure\cite{wang2020deep}, normalized cross-correlation, and bert score\cite{bert-score}. The detailed information is then combined with the task query and input into the MLLM to generate results.}\label{example_figure}
\end{figure*}

\section{Numerical Results}\label{sec:numerical results}

\begin{table}[t]
	\renewcommand{\arraystretch}{1.3}
        \captionsetup{font = small}
	\caption{ Key Hyperparameters in the Simulation. }\label{table} \centering 
	\begin{tabular}{m{5.3cm}<{\raggedright}|m{2.1cm}<{\centering}}	 	
        \toprule[1.5pt]
		\hline		
		\textbf{Hyperparameters} & \textbf{Setting}\\	
		\hline
		Learning rate of the contract generation network &  $1\times10^{-6}$\\	
		\hline
		Learning rate of the contract quality network &  $1\times10^{-6}$  \\	
		\hline
		Soft target update parameter $\tau$  & $0.005$  \\
		\hline
		Exploration noise $\varepsilon$ &  $0.01$  \\	
		\hline		
		Batch Size $N$ &  $512$\\	
		\hline		
		Denoising steps for the diffusion model $T$ &  $5$\\
		\hline
        Maximum capacity of the replay buffer $|\mathcal{D}|$ & $10^{6}$\\
        \hline
        \bottomrule[1.5pt]
	\end{tabular}\label{table_parameter}
\end{table}

In this section, we present extensive experiments to evaluate the performance of the proposed hybrid RAG-empowered MLLM framework for healthcare analysis, as well as the effectiveness of the proposed incentive mechanism. MLLM inference is conducted using Python 3.10.14 on an Intel Xeon(R) Gold 6133 CPU and an NVIDIA RTX A6000 GPU. For the implementation of GDM-based DRL algorithms, the primary parameter settings are detailed in Table \ref{table_parameter}, with experiments run on an NVIDIA GeForce RTX A6000 server GPU using CUDA 11.8.

\subsection{Case Study of Hybrid RAG-empowered Medical MLLMs}
\begin{figure}[t]
\centering{\includegraphics[width=0.48\textwidth]{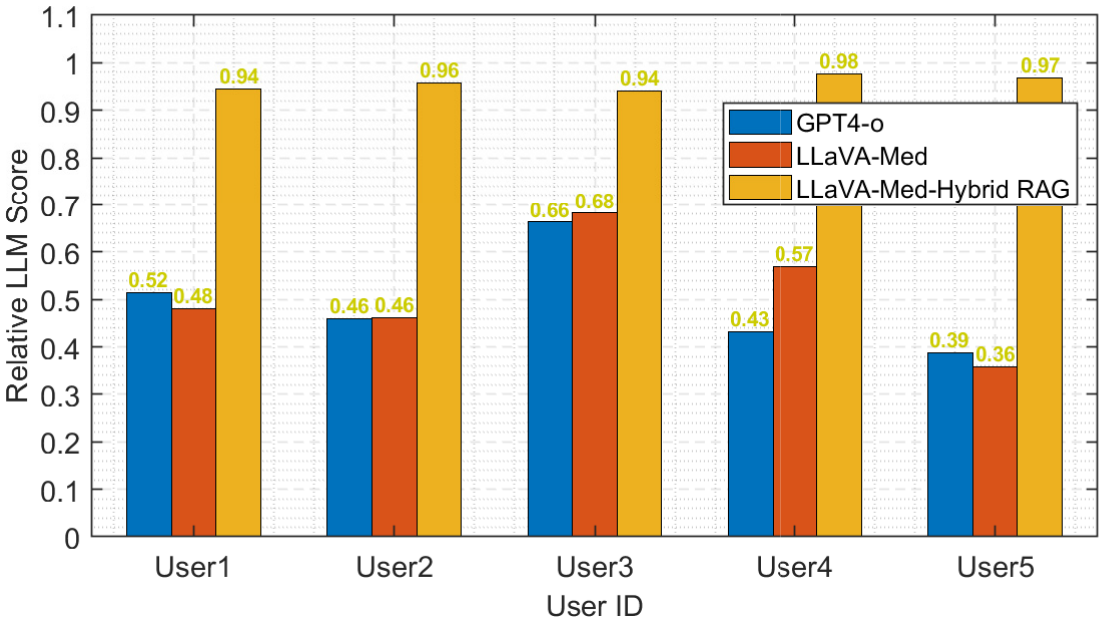}}
\captionsetup{font = small}
\caption{Performance comparison between the proposed framework under different healthcare data cases. Note that the initial two users provide conditional healthcare data cases, while the subsequent three users provide normal healthcare data cases.}\label{LLava_Evaluation}
\end{figure}

\begin{table}[htbp]
\renewcommand{\arraystretch}{1.6}
 \centering
 \tiny
\captionsetup{font = small}
 \caption{The Performance of Different Methods.}
 \setlength{\tabcolsep}{5pt}
 {\fontsize{9}{9}\selectfont
  \begin{tabular}{cccc}
    \toprule[1.5pt]
    \hline
    Methods    & RAI  & SS &  Relative LLM scores  \\\midrule
    GPT4-o     & 0.55 & 0.43 & 0.49 \\
    LLAVA-Med  & 0.54 & 0.48 & 0.51 \\
    \textbf{LLAVA-Med-Hybrid RAG} & \textbf{0.98} & \textbf{0.93} & \textbf{0.96}
    \\\hline
    \bottomrule[1.5pt]
 \end{tabular}
 \selectfont
 }
 \label{The performance of different methods}
\end{table}

We simulate a prototype of the hybrid RAG-empowered MMLMs with the support of LLaVA-Med\cite{li2024llava} and llamaindex\footnote{\url{https://docs.llamaindex.ai/}}. As illustrated in Fig. \ref{example_figure}, we present two examples to demonstrate the functionality and application of our framework. Upon receiving a task query with multi-modal healthcare data, the framework first retrieves the corresponding data from the respective modal database and performs a preliminary screening of $K$ results based on the cosine similarity between vectors. Next, hybrid RAG further refines these results using the MIS metric to identify the best matches, which are then used as inputs to MLLMs. Finally, the MLLMs process all this information to provide diagnostic outputs and personalized services according to the query.

We apply the criteria of Responsive Artificial Intelligence (RAI)\footnote{\url{https://www.microsoft.com/en-us/ai/responsible-ai}} to assess whether the outputs of MLLMs present potential risks related to morality, bias, and ethics. Additionally, we also assess the relationship between the output of MLLMs and task query with the Semantic Similarity (SS)\cite{chandrasekaran2021evolution}, which reflects the diagnosis quality and serves as a crucial indicator for measuring the output of MLLMs. Due to the lack of evaluation benchmarks, we integrate LLM evaluators\cite{wang2023chatgpt} with prompt engineering techniques \cite{liu2023optimizing} to measure the quality of data analysis for MLLMs under the method of GPT4-o, LLaVA, and LLaVA with hybrid RAG. The scoring is normalized to a range of $[0, 1]$. Higher scores denote greater reliability and lack of bias in the MLLM output, along with a strong correlation to task query information. In contrast, lower scores signify a substantial gap from the anticipated results. Finally, we combine the RAI evaluation and SS into a unified score, known as the relative LLM score $\zeta$\cite{huang2024large}, which is calculated using the formula:
\begin{equation}
       \zeta = \lambda \cdot RAI + \nu \cdot SS,
\end{equation}
where $\lambda$ and $\nu$ are the weighting factors for RAI and SS, respectively. In our approach, we assign equal weights by setting $\lambda = 0.5$ and $\nu = 0.5$.

As shown in Fig. \ref{LLava_Evaluation}, we present the performance comparison between the proposed framework under different healthcare data cases. Our findings indicate that hybrid RAG enables LLaVA-Med to consistently score above $0.9$, particularly in X-ray cases from Users 1 and 2 with known etiologies, maintaining high-quality answers and stability. In contrast, other MLLMs exhibit reduced output quality due to the interference of disease factors. In the scenarios involving Users 3 and 4, who are normal without specific causes, MLLMs achieve high scores and deliver reasonable judgments. However, in the case of User 5, who is normal but has an X-ray that can easily be misjudged by a doctor, other MLLMs exhibit a higher misjudgment rate. In contrast, hybrid RAG continues to produce high-quality outputs by matching similar disease conditions. These cases illustrate that the retrieved data retrieved by hybrid RAG provides valuable information for answering questions. We summarize all scores in Table \ref{The performance of different methods}, which clearly demonstrates that hybrid RAG helps LLAVA-Med maintain consistently high scores, showcasing its strong performance across different scenarios. This indicates that hybrid RAG effectively considers the quality of retrieved information by utilizing the features of multi-modal data, including images and texts. The retrieved relevant healthcare data can aid MLLMs through contextual relationships, allowing MLLMs to deliver reliable and robust outputs owing to their powerful contextual learning capabilities.

\subsection{Performance of GDM-based Contract Theory Approach}
\begin{figure}[t]
\centering{\includegraphics[width=0.45\textwidth]{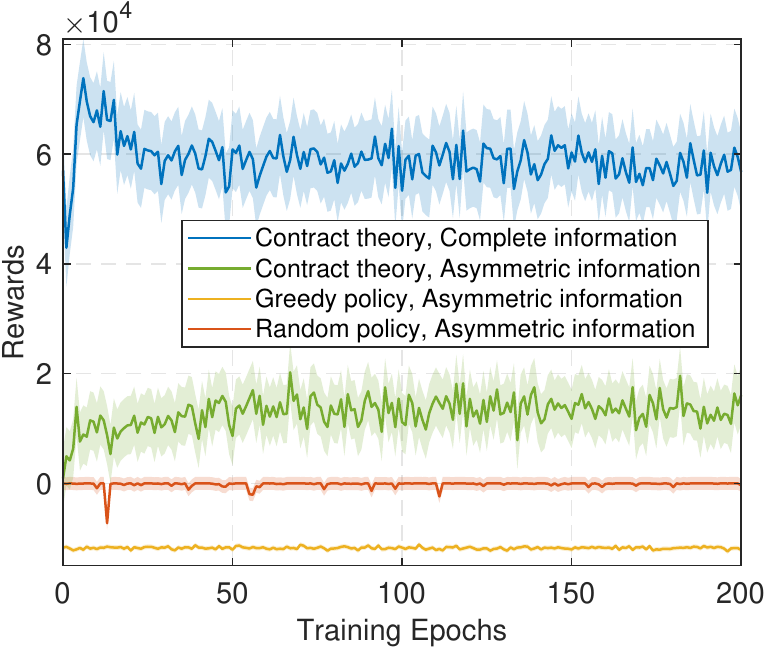}}
\captionsetup{font = small}
\caption{Reward comparison of our scheme with other schemes, i.e., contract-based incentive mechanism with complete information, greedy, and random.}\label{GDM_PPO_information}
\end{figure}

\begin{figure}[t]
\centering{\includegraphics[width=0.45\textwidth]{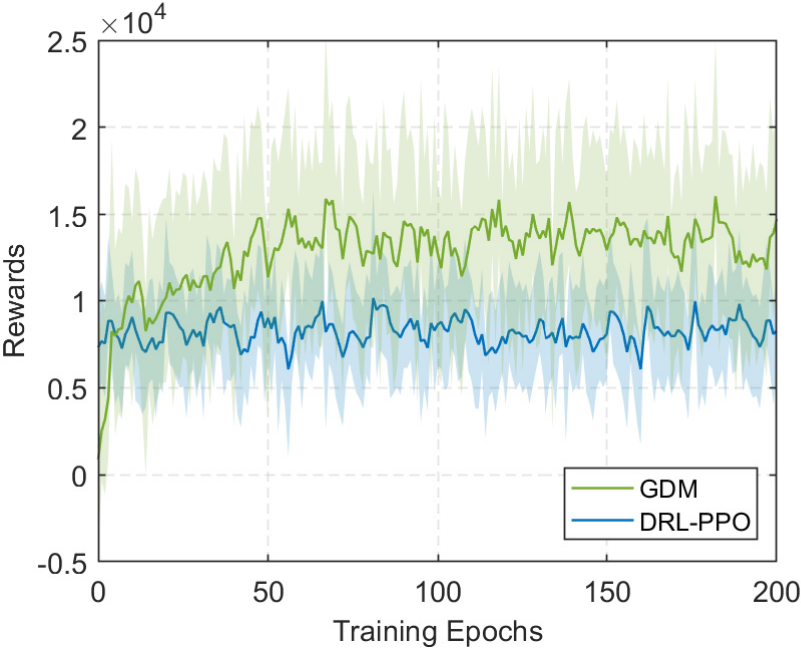}}
\captionsetup{font = small}
\caption{ Performance comparison between the GDM and DRL-PPO in optimal contract design.}\label{GDM_PPO}
\end{figure}

\begin{figure}[t]
\centering{\includegraphics[width=0.45\textwidth]{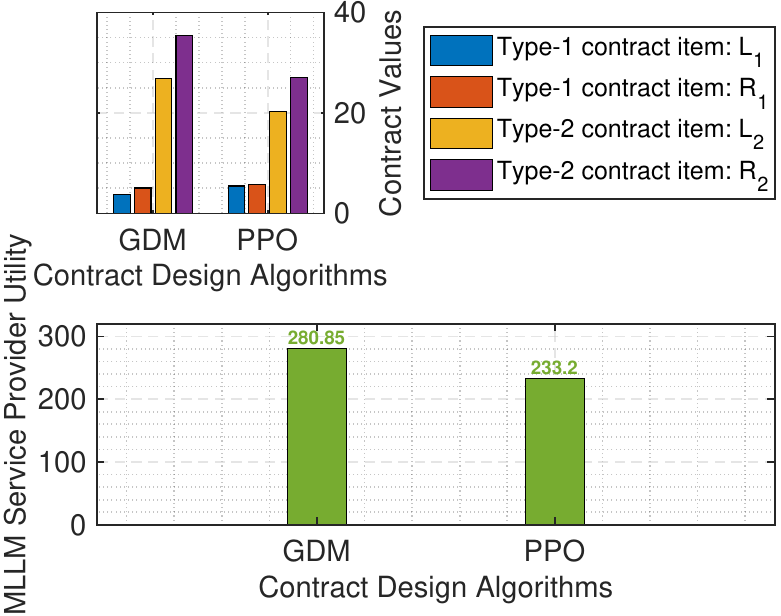}}
\captionsetup{font = small}
\caption{ Optimal contracts designed by the GDM and DRL-PPO.}\label{GDM_PPO_type}
\end{figure}

In the proposed contract model, we employ an on-policy GDM algorithm within a double actor-critic framework for optimal contract design, and the specific settings of training hyperparameters are shown in Table \ref{table_parameter}. In our setup, we consider $10$ healthcare data holders divided into two types, with $M = 10$ and $K = 2$. For the two types of healthcare data holders $\theta_1$ and $\theta_2$, values are randomly sampled from the intervals $[1, 6]$ and $[13, 18]$, respectively. Additionally, the maximum tolerance of AoI $\overline{A}_{max}$ is sampled randomly within the range of $[30, 60]$. For the utility of the MLLM service provider, the parameters $\alpha$, $\beta$, $t$ are set to $39.9$, $10$, and $2$, respectively, and $Q_1$ and $Q_2$ are randomly generated according to the Dirichlet distribution\cite{wen2024IoT}. 

Firstly, we compare our proposed contract-based incentive mechanism, which operates under information asymmetry, with other methods: a contract-based mechanism with complete information, a greedy scheme, and a random scheme. As illustrated in Fig. \ref{GDM_PPO_information}, we can find that our proposed contract scheme consistently outperforms the greedy and random schemes. However, for identical parameter settings, the contract-based mechanism with complete information yields higher performance than our model. This result highlights the disadvantage of information asymmetry, as the MLLM service provider gains fewer benefits without precise knowledge of the types of healthcare data holders. Although a complete information scenario allows the MLLM service provider to offer the optimal contract items to healthcare data holders by knowing their exact types, it is not a realistic environment. In practice, even with complete information, a rational healthcare data holder may provide misleading information to manipulate rewards, ultimately reducing the subjective utility for the MLLM service provider. Thus, our proposed contract model, which handles asymmetric information, proves to be more reliable and practical, achieving the highest utility in real-world scenarios.

In Fig. \ref{GDM_PPO}, we compare the performance of the GDM and DRL with Proximal Policy Optimization (DRL-PPO) in optimal contract design. Both models are capable of continuously acquiring rewards in complex and variable environments until convergence. Notably, the final test rewards of GDMs are significantly higher than those for DRL-PPO under identical parameter settings, allowing the MLLM service provider to consistently secure greater utilities. This is attributed to the fine-grained policy adjustments during the diffusion process, which effectively reduces the impact of randomness and noise\cite{wen2024IoT}. Additionally, exploration through diffusion enhances the flexibility and robustness of the contract design policy, preventing it from falling into suboptimal solutions. Consequently, this superior performance demonstrates the ability of GDMs to capture intricate patterns and connections among environmental observations, and it can effectively reduce the complexity of the relationship between healthcare data holders and the MLLM service provider. 

In Fig. \ref{GDM_PPO_type}, we present the optimal contracts designed by the GDM and DRL-PPO. Given the environmental state, the GDM-based model, enhanced by exploration during the denoising process, produces a contract design that delivers a utility value of $280.85$ for the MLLM service provider, which is higher than the $233.2$ achieved by DRL-PPO. This advantage arises from GDM’s capability to generate near-optimal contracts. Additionally, as the type of healthcare data holder increases, the rewards they receive also rise. However, DRL-PPO shows consistent variables for the type-$1$ healthcare data holders, indicating a tendency towards local optimal solutions, which may not align with global interests. Overall, this numerical analysis highlights the practical feasibility and superior performance of the proposed GDM-based scheme.

\subsection{Secure Block Verification Performance Analysis}

To assess the security of the blockchain system, we evaluate the reputation value of hospitals. Each hospital's associated subchain generates a block and broadcasts it to the relay chain for validation. If validated, the relay chain submits the block to the main chain linked to the health center. The health center then rewards each hospital according to their actions, applying a reputation-based bonus and penalty system.

\begin{figure}[t]
\centering{\includegraphics[width=0.45\textwidth]{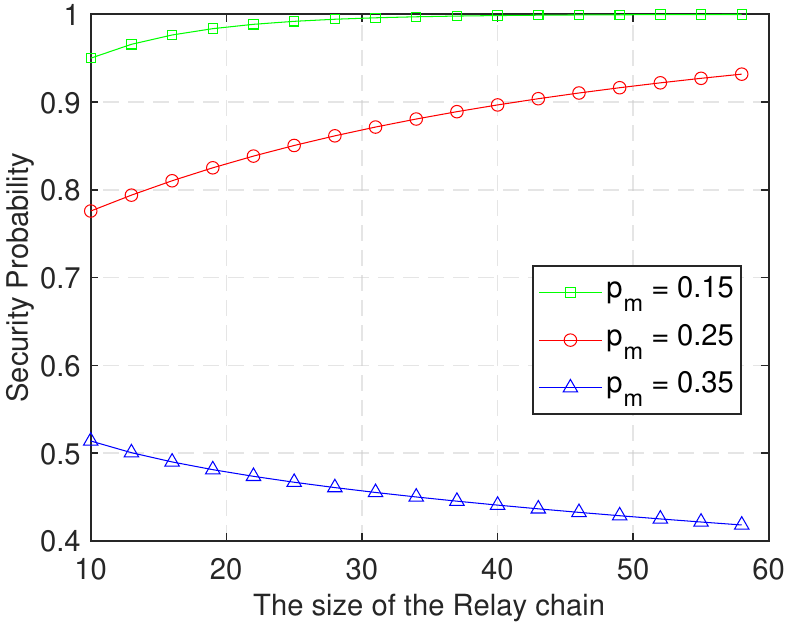}}
\captionsetup{font = small}
\caption{Security probability at varying malicious miner probabilities.}\label{Blockchain_Security}
\end{figure}

As illustrated in Fig. \ref{Blockchain_Security}, we use the Practical ByzantineFault Tolerance (PBFT) consensus algorithm to assess the security performance of the blockchain system. We consider that the subchains operate reliably and model the relay chain's security performance as a random sampling problem with two potential outcomes, i.e., malicious delegates and well-behaved delegates\cite{zhong2023blockchain}. When the number of malicious delegates is no greater than $(N - 1)/3$, where $N$ represents the total number of delegates, the block verification process remains accurate\cite{zhong2023blockchain}. Therefore, the probability of secure consensus, denoted as $P_{\mathrm{safety}}=\sum_{z=0}^{\lfloor N/3\rfloor}\binom Nzp_m^z(1-p_m)^{N-z}$, depends on $p_m$ that represents the probability of a delegate being malicious. Figure \ref{Blockchain_Security} shows that as the size of the relay chain increases, the security probability also rises, regardless of the likelihood of malicious delegates. This improvement is due to the larger number of well-behaved delegates involved in block validation, which strengthens security in the consensus process. Thus, the proposed blockchain system with the PBFT consensus algorithm supports secure and reliable data sharing by ensuring robust block verification.

\section{Conclusion}\label{sec:Conclusion}
In this paper, we have studied the service quality issues of MLLMs and the design of incentive mechanisms for healthcare data management. We have proposed a hybrid RAG-empowered medical MLLM framework based on cross-chain technologies to enhance healthcare data management in IoMT. Specifically, we have utilized a cross-chain structure comprising a main chain and multiple subchains to ensure the security of healthcare data. Additionally, we have applied hybrid RAG with multi-modal information similarity metrics to retrieve similar healthcare data, thereby improving the quality of MLLM services. Then, we have applied AoI to indirectly quantify the quality of healthcare data and utilized contract theory to incentivize healthcare data holders to contribute high-quality healthcare data with small AoI, thus enhancing the quality of MLLM services. Furthermore, we have employed GDMs to generate the optimal contracts for efficient data sharing. Finally, numerical results demonstrate the effectiveness and reliability of our proposed framework and incentive mechanism. For future work, we aim to enhance our framework's performance by integrating additional characteristics of multi-modal healthcare data and developing a multi-dimensional contract model to better address the complexities of IoMT environments.


\bibliographystyle{IEEEtran}
\bibliography{main}

\end{document}